%% file: root.tex
\documentclass[letterpaper, 10 pt, conference]{ieeeconf}
\IEEEoverridecommandlockouts
\usepackage{graphicx} 
\graphicspath{{figures/}}
\usepackage{amsmath} 
\usepackage{amssymb}  
\usepackage{algorithm}
\usepackage{algorithmic}
\usepackage{hyperref}
\usepackage{tabularx}
\usepackage{booktabs}
\usepackage{array}
\usepackage{ragged2e}
\usepackage{comment}
\usepackage{xcolor}
\usepackage{censor}
\usepackage{caption} 
\usepackage{subcaption}
\usepackage{multirow}
\title{\LARGE \bf
Personalized and Robust Proactive Robot Assistance 
with Uncertainty-Guided LLM Reasoning
}

\author{
Alvaro Gonzalez$^{1}$, M.H.~Hasan~Shovo$^{1*}$, Ali~Ayub$^{1}$
\thanks{
This research was undertaken, in part, thanks to funding from the Natural Sciences and Engineering Research Council of Canada (NSERC) and Concordia University.}
\thanks{$^{1}$Concordia University, Montreal, Quebec H3G 1M8, Canada}
\thanks{{\tt\small a\_zalezl@live.concordia.ca, \{$*$mdhasibulhasan.shovo, ali.ayub\}@concordia.ca}}
}

\begin{document}

\maketitle
\thispagestyle{empty}

\input{sections/abstract}

\input{sections/introduction}

\input{sections/related_work}

\input{sections/methodology}

\input{sections/experiments}

\input{sections/conclusion}

\bibliographystyle{IEEEtran}
\bibliography{IEEEabrv, references}

\end{document}

%% file: sections/abstract.tex
\begin{abstract}
Proactive robot assistance in household environments requires accurate prediction of human activities and object usage under dynamic and noisy conditions. Existing approaches often rely on complex spatio-temporal models, which can be computationally expensive and sensitive to environmental variability. In this paper, we propose GLOBE, a lightweight framework that combines n-gram Markov models for capturing temporal behavioral patterns with uncertainty-guided large language model (LLM) reasoning. The framework performs sequential prediction efficiently while selectively invoking LLM reasoning only when the model confidence is low.
To evaluate performance under realistic conditions, we introduce HOMER-Noise, a noisy extension of the HOMER+ dataset that simulates structured disturbances such as object movements caused by humans, pets, and toddlers. Experimental results show that GLOBE achieves competitive performance with state-of-the-art methods while improving robustness and computational efficiency across both clean and noisy settings. The framework is further validated through a proof-of-concept integration with a Stretch 3 mobile manipulator, demonstrating its potential application in real-world human–robot interaction scenarios. Code is available at  \href{https://github.com/PaInt-Lab/GLOBE}{\texttt{https://github.com/PaInt-Lab/GLOBE}}
\end{abstract}


%% file: sections/introduction.tex
\section{Introduction}
\label{sec:introduction}

\noindent
Robots deployed in everyday human environments are expected to assist users in a proactive and seamless manner, reducing the need for explicit user commands. Therefore, proactive assistance has been identified as a key principle for effective human–robot interaction (HRI), improving both usability and user experience in long-term deployments \cite{
gross_robot_2015}. In household settings, this requires the robot to anticipate user needs by predicting future activities and the objects required to perform them. Prior work has studied assistance at two time scales: short-term assistance conditioned on the user’s current action, such as handing over the next tool during an assembly task~\cite{puig2021watchandhelp
}, and longitudinal assistance, in which the robot anticipates the user’s needs over longer horizons, such as setting up a table for breakfast before the user enters the kitchen~\cite{patel2023predicting}. In this paper, we address the problem of personalized longitudinal proactive robot assistance for household tasks. 

\begin{figure*}[t]
\centering
\includegraphics[width=.75\linewidth]{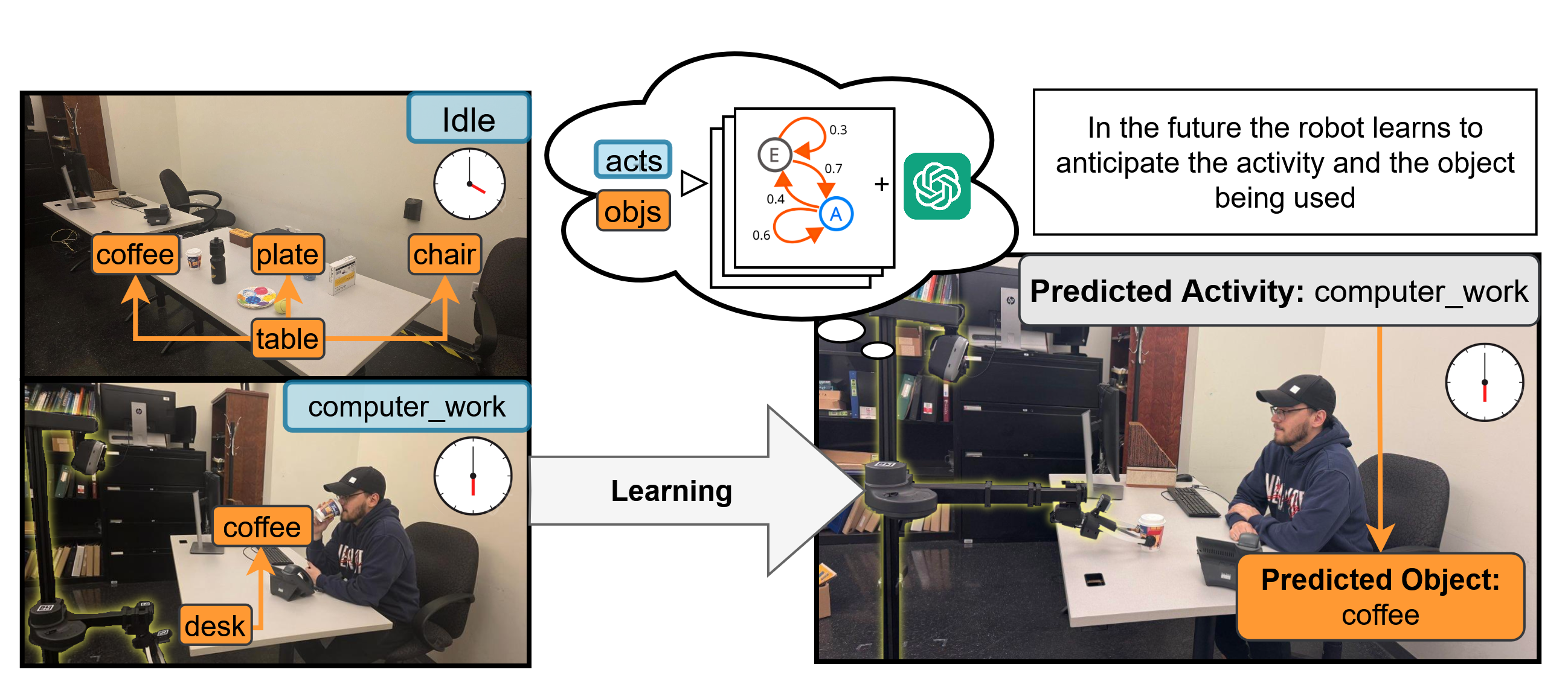}
\caption{
\textbf{Left:} The robot observes the environment and user activity (e.g., idle, computer\_work), identifying relevant objects and their spatial context. \textbf{Middle:} The system learns temporal relationships between activities and object usage using sequential modeling, augmented with LLM 
reasoning for handling uncertainty. \textbf{Right:} Given the predicted activity, 
the framework predicts the most likely object to be used next (e.g., coffee), enabling proactive robot assistance
.}
\label{fig:overview_framework}
\end{figure*}

Recent work has explored spatio-temporal modeling techniques to predict object usage for this problem. Early approaches, such as STOT~\cite{patel2022proactive} model object dynamics using structured representations of the environment and temporal patterns
. Building on this, SLaTe-PRO~\cite{patel2023predicting} 
incorporates activity information and latent temporal representations to improve long-horizon predictions 
using transformers, graph neural networks, and feedforward multi-layer perceptrons (MLPs). Similarly, advances in activity recognition leverage deep neural networks to infer human actions from sensor data \cite{
alaghbari_activities_2022}. Despite this progress, these methods remain sensitive to noise and activity variability commonly observed in real-world environments. 
Moreover, these methods primarily focus on improving predictive accuracy through increasingly sophisticated models, with limited emphasis on efficiency, robustness, and adaptability, which are critical for real-world HRI systems \cite{ayub2024interactive, ayub_continual_2025}. 

In this paper, we address 
these gaps 
and propose a lightweight framework, uncertainty-Guided LLM reasoning for routine Object prediction via Behavioral tEmporal modeling (GLOBE), to predict 
object usage for proactive robot assistance in household environments. To provide an intuitive understanding of the proposed approach, Fig.~\ref{fig:overview_framework} illustrates how the framework operates in a real-world scenario using a Stretch 3 mobile manipulator. Our framework trains n-gram Markov models~\cite{markov_example_2006} to capture temporal patterns in user activity and object movement behavior. When the Markov model object predictions are uncertain, the framework actively prompts a pre-trained large language model (LLM) to utilize it's semantic reasoning ability for reliable object prediction.
This design enables efficient sequential prediction while selectively leveraging LLM reasoning when required, improving robustness to noise and variability without increasing model complexity. We evaluate our method on the noise-free HOMER+~\cite{patel2023predicting} dataset and compare it against state-of-the-art (SOTA) approaches
. To further assess real-world applicability, we introduce \emph{HOMER-Noise}, a noisy extension of HOMER+ that simulates realistic environmental disturbances, such as 
object movements introduced by humans \& pets, in the HOMER$+$ dataset. Our results show that the proposed framework achieves competitive predictive performance while significantly improving robustness and 
training efficiency on both HOMER+ and HOMER-Noise datasets. 

%% file: sections/related_work.tex
\section{Related Work}
\label{sec:related_work}

\noindent
Proactive assistance in HRI has been widely studied as a key capability to enable robots to anticipate and support human activities in collaborative environments. Early works focused on predicting human actions and intentions to facilitate coordination, such as anticipating human motion for collaborative manipulation \cite{mainprice_human-robot_2013} and reasoning under task and sensor uncertainty \cite{hawkins_anticipating_2014}. These approaches demonstrate the importance of early prediction for effective collaboration but are typically limited to short-horizon tasks and structured settings. Other works incorporate activity recognition and human instruction to enable teamwork between humans and robots \cite{cuntoor_human-robot_2012}, highlighting the role of semantic understanding in assistive behaviors. More recent systems extend these ideas to long-term and personalized interactions, emphasizing adaptability and user-centric assistance in domestic environments \cite{ayub2024interactive, ayub_continual_2025}.
While these approaches focus on predicting human actions or enabling collaboration, they do not explicitly model the relationship between activities and object usage required for proactive assistance over longer time horizons.

To bridge this gap, STOT~\cite{patel2022proactive}, and its extension SLaTe-PRO~\cite{patel2023predicting}, model object dynamics using structured representations of the environment and temporal dependencies, achieving strong predictive performance in household settings. In parallel, deep learning activity recognition methods have demonstrated success in inferring human actions from sensor data in smart environments \cite{
alaghbari_activities_2022}.
Despite improved prediction accuracy, the performance of these models can be sensitive to variations in activity patterns and environmental conditions, which are common in real-world settings.

More recently, large language models (LLMs) have emerged as a promising tool for reasoning and decision-making in robotics. LaMI \cite{wang_lami_2024} and AssistantX \cite{sun_assistantx_2025} demonstrate how LLMs can be integrated into HRI systems to enable multi-modal interaction and proactive assistance, while other approaches explore aligning uncertainty in LLM-based planners for interactive decision-making \cite{ren_robots_2023}.
However, using LLMs as sole reasoning models can introduce latency, cost, and reliability concerns. In contrast, our work adopts a hybrid approach that combines lightweight sequential modeling with selective LLM-based reasoning, aiming to balance efficiency, robustness, and adaptability for proactive object usage prediction in real-world environments.

%% file: sections/methodology.tex
\section{Methodology}
\label{sec:methodology}

\subsection{Problem Formulation}

\noindent
We consider a household environment consisting of a set of objects \( O = \{o_i\} \) and locations \( L = \{l_j\} \)
. The state of the environment at time \( t \) is represented as
\(
G_t = \{(o_i, l_j)\},
\)
which defines the location of each object in the environment and is assumed to be observable by the robot. Let \( A \) denote a predefined set of human activities. 

We formulate proactive assistance as a two‑phase process consisting of an observation (training) phase and an assistance (inference) phase. During the observation phase, the robot receives sequential observations of the environment state $G_t$ and user activities $a_t \in A$, which it uses to learn a predictive model of the user’s behavior. To reduce ambiguity in object prediction, the system also associates each activity with a set of objects based on co-occurrence patterns observed during this phase. For a given activity \(a_t\), we define \(O(a_t) \subseteq O\) as the set of objects that are commonly used during that activity.
In the assistance phase, given a history of observed environment states \( G_{0:t} \) and past activities \( a_{0:t-1} \), the 
system first predicts the current activity \( \hat{a}_t \), and then predicts the object \( \hat{o}_t \in O(\hat{a}_t) \) that is most likely to be used at time \( t \).
The prediction task is then defined as:
\(
\hat{o}_t = \arg\max_{o \in O(\hat{a}_t)} p(o \mid a_{0:t-1}, G_{0:t}),
\)
where the objective is to select the most likely object given the available context.

\begin{figure}[!t]
\centering
\includegraphics[width=\linewidth]{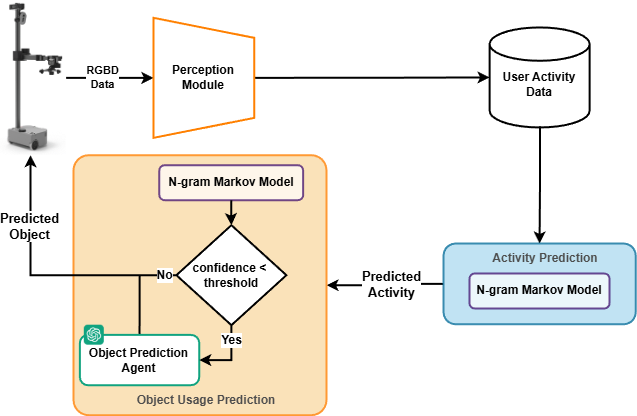}
\caption{Overview of the proposed framework for object usage prediction. The system learns from user activity data and, during the inference phase, predicts the next activity using an n-gram Markov model. The predicted activity is then used to guide object prediction through a sequential model, augmented with uncertainty-guided LLM reasoning. 
The final predicted object is 
sent to the robot for manipulation.}
\label{fig:framework}
\vspace{-1.5em}
\end{figure}

\subsection{Sequential Prediction Framework}
\label{sec:seq_pred_framework}


\noindent
Fig.~\ref{fig:framework} provides an overview of our reasoning framework. It operates in a perception-driven loop, where the model first predicts the next activity, based on knowledge gained in the observation (training) phase, followed by object usage prediction conditioned on the predicted activity.

\subsubsection{Activity Prediction}
\label{sec:activity_Pred}

During the observation phase, the robot records the user’s activities as ordered daily sequences \( a_1, a_2, \dots, a_t \), where each \( a_t \in A \). These sequences are used to compute frequency counts of activity transitions for different context lengths, which form the basis of the n-gram Markov model \cite{markov_example_2006}.
At time step \( t \), the model uses the recent activity history as context, defined as \( c_{act} = (a_{t-n}, \dots, a_{t-1}) \). The probability of the next activity is estimated as:
\[
P(\hat{a}_t \mid c_{act}) = \frac{\text{count}(c_{act} \rightarrow a_t)}{\sum_{a \in A} \text{count}(c_{act} \rightarrow a)}
\]

During inference, the available history length varies across time steps. The model, therefore, first leverages the longest observed activity sequence to predict the next activity. 
If this sequence was not encountered during training, progressively shorter recent sequences are used to estimate the transition probability. 
This design enables the model to exploit fine-grained temporal structure when sufficient context is available, while remaining effective under limited observation.
This formulation also improves robustness to noisy activity observations. Because predictions are derived from aggregated transition statistics over multiple sequence lengths, infrequent 
transitions have limited influence. When unseen or unreliable patterns occur, the model naturally backs off to shorter, more frequently observed sequences, yielding more stable and reliable predictions.

\subsubsection{Object Prediction}

Object prediction is performed after predicting the activity \( a_t \), following the sequential formulation described in Section~\ref{sec:activity_Pred}. 
 During observation, the robot records object usage sequences conditioned on activity, maintaining separate transition counts for each~\( a_t \in A \) 
to capture activity-specific object dynamics.
At time step \( t \), object prediction is conditioned on the predicted activity \( a_t \) 
and recent object history within that activity, defined as \( c_{obj} = (o_{t-n}, \dots, o_{t-1}) \), where \( o_{t-1} \) represents the most recently used object. The probability of the next object is estimated as:
\[
P(\hat{o}_t \mid c_{obj}, a_t) = \frac{\text{count}(c_{obj} \rightarrow o_t)}{\sum_{o \in O(a_t)} \text{count}(c_{obj} \rightarrow o)}
\]

Following the formulation in Section~\ref{sec:methodology}, object prediction is restricted to the set \( O(a_t) \), 
associated with the activity \( a_t \), ensuring consistency with the current activity context. During inference, the model uses the longest available object history; if an exact sequence was not observed during training, it backs off to shorter histories. This allows reliable prediction under sparse or noisy object usage patterns.

\subsection{LLM-based Adaptive Reasoning}

\noindent
While sequential models are effective at capturing recurring patterns, they can struggle 
when the context is sparse, ambiguous, or previously unseen. To address this
, we incorporate an LLM as 
a semantic reasoning module that is selectively invoked under uncertainty.
When the 
sequential model prediction confidence falls below a predefined threshold ($d$, a hyperparameter), the system queries the LLM to refine the prediction. The LLM is provided with 
the recent activity, object sequence, and a set of top-$k$ predictions obtained from the n-gram model. Based on this information, it produces a refined prediction that accounts for broader contextual knowledge and semantic relationships.
The following example illustrates next-object prediction using the LLM (the LLM response is \colorbox{teal!20}{highlighted}):

\fbox{\parbox{0.9\linewidth}{
\texttt{Given activity ``breakfast'' and this object sequence:}\\
\texttt{spoon -> bowl -> milk}\\
\texttt{what is the most likely next moved object? 
Return only one object name from this list:}\\
\texttt{[cup, plate, cereal\_box, knife]}

\# LLM Response: \colorbox{teal!20}{\texttt{cup}}}
}

\subsection{HOMER-Noise}
\noindent
We contribute HOMER-Noise, an extension of HOMER+ \cite{patel2023predicting} designed to capture environmental disturbances in simulated households. 

\subsubsection{Motivation and Limitations of HOMER+}

HOMER+ consists of week-long sequences of household activities and object-location states, capturing temporal variability in activity timing as well as variation in object usage within activities. These characteristics make it suitable for predicting activities and object usage in domestic settings. However, HOMER+ assumes that changes in the household arise primarily from the executed activities, and does not explicitly model external environmental noise
. 
In real households, objects may be moved unintentionally, reorganized, or displaced by interactions unrelated to the current activity. For instance, while eating breakfast, a user may move a remote control from the table to the sofa for tidying or other incidental reasons. Such behaviors introduce variability not captured in HOMER+, limiting the realism of the object-prediction task.

\subsubsection{Creating HOMER-Noise}
To create HOMER-Noise, we augment HOMER+ with object-location disturbances that represent plausible real-world interactions. 
Rather than introducing arbitrary perturbations, noise is modeled in a semantically grounded manner to ensure consistency with the household environment. 
We consider three noise types. \emph{Person noise} captures object movements 
arising from typical human behaviors, such as reorganizing, tidying, or temporarily relocating items. For example, during a \textit{watch\_tv} activity, a user may move a snack item closer for convenience. \emph{Pet noise} simulates incidental object displacements caused by pets 
, such as a dog pushing a toy or a cup while moving around the room. \emph{Toddler noise} represents less structured and 
more unpredictable interactions
, including randomly moving household objects, such as utensils across locations. 
These disturbances are generated by prompting an LLM with the current activity, object sequence, and environment context, enabling the simulation of plausible object movements under each noise condition. An example of the prompt used to generate such disturbances is shown below (the LLM response is \colorbox{teal!20}{highlighted}):

\fbox{\parbox{0.9\linewidth}{
\texttt{Given a household scenario with activity ``breakfast'' and object sequence:}\\
\texttt{bowl -> spoon -> milk, generate additional object movements caused by a toddler.}\\
\texttt{Ensure the movements are realistic and consistent with a household environment.}

\# LLM Response: 
\colorbox{teal!20}{
\parbox{0.60\linewidth}{
\texttt{toddler moves spoon to \\ floor, moves cup to living\_room table}
}
}
}
}

To control the amount of noise inserted, we vary the number of injected object movements relative to the original total generated amount. For example, a noise level of 0.5 indicates that half of the total generated noise was inserted into the dataset.

\subsubsection{Characteristics of HOMER-Noise}
The resulting dataset preserves the activity labels, temporal structure, and object information of HOMER+ while introducing controlled disturbances into the environment. Object locations are therefore no longer determined solely by scripted activity routines, but may also be influenced by secondary interactions involving a person, pet, or toddler. This increases the difficulty of object prediction, as models must distinguish activity-induced changes from those caused by external noise. By incorporating these perturbations, HOMER-Noise better captures the uncertainty and variability of real-world household environments, providing a more realistic benchmark for evaluating predictive systems. 

%% file: sections/experiments.tex
\section{Experiments}
\label{sec:experiments}

\subsection{Datasets and Pre-processing}
\noindent
We evaluate our model on the HOMER+~\cite{patel2023predicting} and HOMER-Noise datasets to assess performance under both clean and noisy conditions. HOMER+ provides longitudinal household activity and object interaction sequences. The dataset captures realistic temporal patterns and variability in user behavior, making it suitable for evaluating proactive assistance systems.
The raw dataset consists of object-location states recorded over time. During preprocessing, we convert these raw logs into two structured representations: (1) activity sequences, and (2) object movement sequences. Activity data is represented as ordered sequences per day, where each time step corresponds to a discrete activity. Object data is represented by recording only object movements, i.e., changes in object location between consecutive time steps, along with the associated activity and timestamp. This transformation removes redundant static information and focuses on meaningful object interactions, making it suitable for sequential modeling. For object prediction, object movement sequences are further organized by activity, enabling the model to learn activity-specific object usage patterns. Objects are filtered based on their observed co-occurrence with each activity in the training set, ensuring consistency between predicted objects and the underlying activity context.
To evaluate robustness, we additionally report results on HOMER-Noise with varying noise magnitudes, as described in subsequent subsections.



\subsection{Evaluation Framework}
\noindent
We evaluate the proposed framework across multiple dimensions, including overall predictive performance, activity \& object prediction performance, robustness under structured noise, and computational efficiency. 
Performance is assessed at each discrete time step \( t \) by comparing the predicted activity and object \( \hat{a}_t, \hat{o}_t \) with the ground-truth \( a_t, o_t \). To evaluate robustness under realistic conditions, we generate noisy variants of the dataset by injecting structured perturbations as described in Section~\ref{sec:methodology}. During training, we apply a moderate noise level of 0.5, while testing is conducted under a higher noise level of 1.0 to assess generalization to more challenging environments. Our primary evaluation metric is the mean F1-score over the joint activity–object output, computed by combining activity and object labels into a unified label space and averaging F1-scores across all activity–object pairs. We additionally report separate F1-scores for activity and object prediction to provide fine-grained insight into model performance.

\begin{table}[t]
\centering
\small
\setlength{\tabcolsep}{6pt}
\renewcommand{\arraystretch}{1.15}
\caption{Performance comparison of different methods on the HOMER+ dataset in terms of F1-score for activity prediction, object prediction, and their mean.}
\label{tab:homer_base_performance}
\begin{tabular}{lccc}
\toprule
\textbf{Method} & \textbf{Activity (F1)} & \textbf{Object (F1)} & \textbf{Mean (F1)} \\
\midrule
\textbf{GLOBE (Ours)} & \textbf{0.80} & 0.65 & 0.73 \\
SLaTe-PRO & 0.78 & \textbf{0.69} & \textbf{0.74} \\
STOT & -- & -- & 0.43 \\
n-gram Markov & 0.79 & 0.64 & 0.72 \\
LLM & 0.57 & 0.00 & 0.29 \\
\bottomrule
\end{tabular}
\vspace{-.5em}
\end{table}

\subsection{Baselines}
\noindent
We evaluate the proposed framework against both classical and state-of-the-art (SOTA) approaches for object usage prediction. Our primary baseline is SLaTe-PRO \cite{patel2023predicting}, which models object dynamics using latent spatial-temporal representations
. We also include STOT \cite{patel2022proactive} 
representing earlier 
work based on structured object-centric modeling. To isolate the contribution of the n-gram Markov model and the LLM, we report results for the Markov model and the LLM evaluated individually.
For the standalone LLM baseline, we use a zero-shot prompting strategy. The LLM receives only the current activity context, the observed sequence, and the candidate prediction set, without access to user-specific training data or learned transition statistics.

\subsection{Implementation Details}
\noindent
We use 65 days of data for training and 10 days for testing, following the standard split used in HOMER+. The maximum n-gram length is set to 109, corresponding to the maximum number of time steps per day. 
The proposed framework is deterministic and based on frequency statistics extracted from training sequences, without requiring gradient-based optimization or parameter tuning.

We set a confidence threshold $d$ of 0.5. When the sequential model assigns a probability below this threshold, indicating insufficient statistical support, the LLM is queried to refine the prediction. 
 In such cases, the LLM is provided with a filtered candidate set consisting of the top-$k$ predictions ($k=5$) from the n-gram model and retaining only those with probability greater than 0.1. The LLM then returns a refined prediction from this candidate set. The LLM is not used for activity prediction. 
We use GPT-4.1 with a temperature of 0.5 and a maximum output length of 10 tokens. All LLM prompts use a zero-shot strategy based on the current activity context and observed sequence, without access to user-specific training histories or in-context examples. 
All experiments are conducted on a system with an NVIDIA RTX 4070 GPU.

\begin{figure*}[!t]
\centering
\includegraphics[width=.85\linewidth]{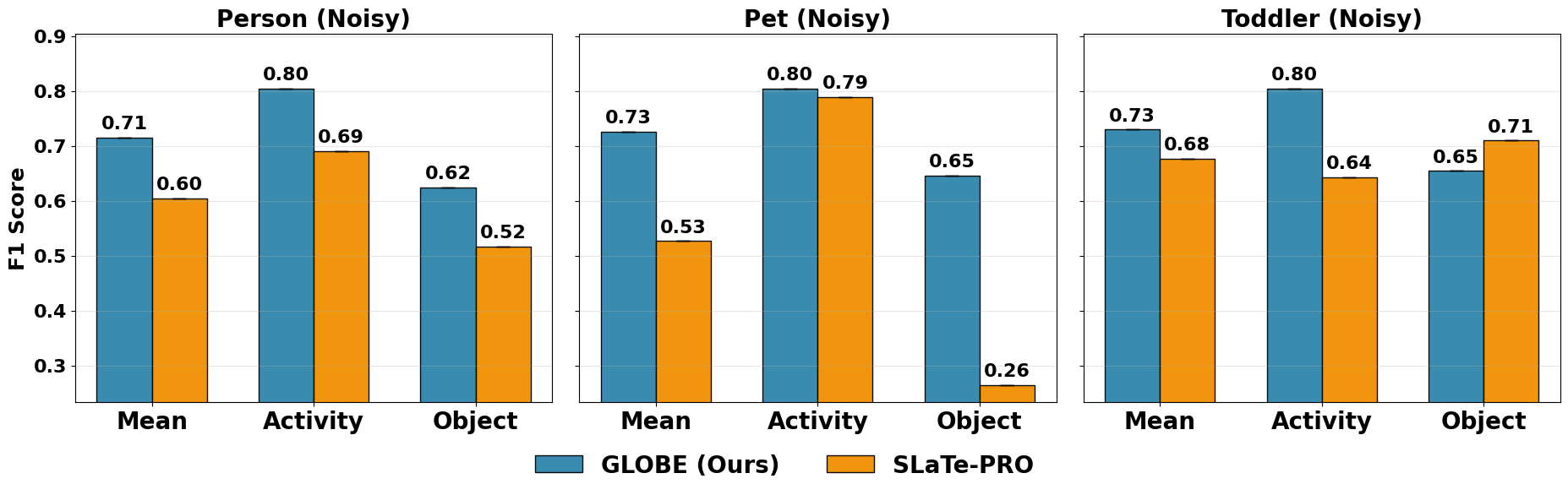}
\caption{Comparison of our model (blue) across different structured noise conditions (person, pet, and toddler) in terms of activity, object and mean F1-score on HOMER-Noise dataset using SLaTe-PRO as the baseline (orange).}
\label{fig:noisy_comparison}
\end{figure*}

\begin{figure}[!t]
\centering
\includegraphics[width=\linewidth]{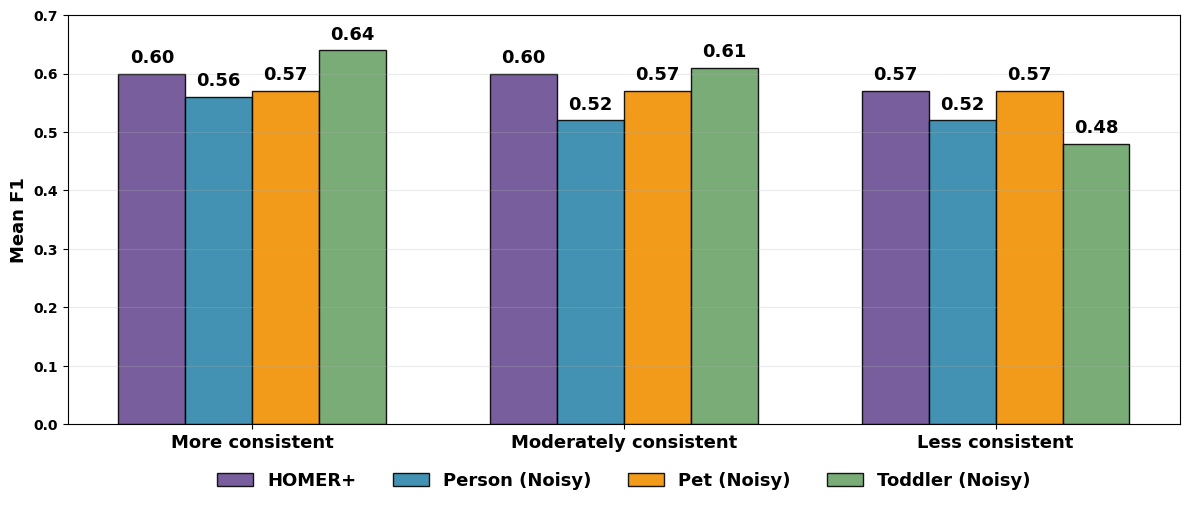}
\caption{Object prediction performance of GLOBE under varying levels of user behavioral consistency on HOMER+ and HOMER-Noise (Person, Pet, Toddler) datasets.}
\label{fig:consistency_analysis}
\vspace{-1.3em}
\end{figure}

\subsection{Results and Analysis}
\subsubsection{Quantitative Performance on HOMER+}

Table~\ref{tab:homer_base_performance} presents the performance of GLOBE and baseline methods on the HOMER+ dataset in terms of activity, object, and mean F1-scores. GLOBE achieves a mean F1-score of 0.73, closely matching the performance of SLaTe-PRO (0.74)
. The results for STOT are taken from the SLaTe-PRO paper~\cite{patel2023predicting}, where it achieves a significantly lower mean F1-score of 0.43. The n-gram baseline performs remarkably well, with an F1 score of 0.72, but the LLM-only baseline achieves a 0.29 F1 score. F1 scores for activity and object predictions provide further insight into these results.

For activity prediction, GLOBE achieves the highest performance with an F1-score of 0.80, slightly outperforming both SLaTe-PRO (0.78) and the n-gram baseline (0.79) and significantly outperforming LLM baseline (0.57), demonstrating its ability to effectively capture temporal activity patterns. For object prediction, GLOBE achieves an F1-score of 0.65, improving over the n-gram baseline (0.64) while remaining competitive with SLaTe-PRO (0.69). LLM baseline does poorly on the object prediction, achieving an F1 score of 0.0. These results demonstrate that GLOBE and the n-gram model perform similarly to more complex SOTA models. The small performance difference between GLOBE and the standalone n-gram model on HOMER+ suggests that sequential patterns alone are sufficient for many routine behaviors. However, the LLM component helps maintain performance under structured noise and uncertain conditions. Additionally, LLM models might not be best suited for zero-shot object prediction, but can assist traditional models in uncertain situations.

\subsubsection{Computational Efficiency}

Table~\ref{tab:method_latency} compares the training and inference time of GLOBE with baseline methods. GLOBE requires only 8.33 seconds for training, which is significantly lower than SLaTe-PRO (11,415 seconds) and comparable to the n-gram baseline. This highlights the ability of our model to quickly adapt to any changes in the real-world environments, enabling lifelong learning. 
During inference, GLOBE incurs higher latency (192.49 seconds) compared to both SLaTe-PRO (93 seconds) and the n-gram baseline (0.86 seconds). This increase is primarily due to the selective use of LLM-based reasoning, where external API calls introduce additional overhead. However, since the LLM is invoked only when the sequential model has low confidence, this cost is incurred selectively and enables improved robustness in challenging scenarios. 

\begin{table}[!t]
\centering
\small
\setlength{\tabcolsep}{6pt}
\renewcommand{\arraystretch}{1.15}
\caption{Comparison of training and inference time across different methods.}
\label{tab:method_latency}
\begin{tabular}{lcc}
\toprule
Method & \textbf{Training (s)} & \textbf{Inference (s)} \\
\midrule
\textbf{GLOBE (Ours)} & \textbf{8.33} & 192.49 \\
SLaTe-PRO & 11,415 & 93 \\
n-gram Markov & 8.45 & \textbf{0.86} \\
\bottomrule
\end{tabular}
\vspace{-1.0em}
\end{table}

\subsubsection{Robustness under Structured Noise}

Figure~\ref{fig:noisy_comparison} compares GLOBE and SLaTe-PRO under different structured noise conditions on the HOMER-Noise dataset.
In terms of mean F1-score, GLOBE consistently outperforms SLaTe-PRO across all noise types, with margins of 0.09, 0.2, and 0.05 for Person, Pet, and Toddler noise conditions, respectively. 
For activity prediction, GLOBE maintains a stable performance of 0.80 across all noise conditions, 
outperforming 
SLaTe-PRO (0.69 for person, 0.79 for pet, and 0.64 for toddler). For object prediction, however, GLOBE outperforms Slate-PRO on Person and Pet noise conditions by margins of 0.1 and 0.39, but underperforms Slate-PRO for the Toddler noise condition with a margin of 0.06. 
The improved performance under structured noise highlights the benefit of the selective LLM reasoning component. While the n-gram model performs strongly in routine settings, the LLM helps recover predictions when noisy object movements create ambiguous or previously unseen contexts.
Overall, GLOBE demonstrates more stable and robust performance across different noise conditions, particularly in object prediction under challenging disturbances, whereas Slate-PRO struggles to handle the noise. 

\subsubsection{Impact of Behavioral Consistency}

Figure~\ref{fig:consistency_analysis} analyzes object prediction performance under different levels of user behavioral consistency across HOMER+ (Original) and HOMER-Noise datasets. Under more consistent behavior, GLOBE achieves an F1-score of 0.60 on HOMER+, compared to 0.56 (person noise), 0.57 (pet noise), and 0.64 (toddler noise). As consistency decreases to a moderate level, performance remains stable at 0.60 on HOMER+, while achieving 0.52 (person), 0.57 (pet), and 0.61 (toddler). Under less consistent behavior, performance slightly drops to 0.57 on HOMER+, with corresponding scores of 0.52 (person), 0.57 (pet), and 0.48 (toddler).
Across all consistency levels, the performance gap between HOMER+ and HOMER-Noise remains moderate, indicating that the model generalizes well under behavioral variability. 
These results demonstrate that GLOBE maintains stable object prediction performance even when user routines become less predictable, highlighting its robustness to both behavioral inconsistency and environmental noise.

\subsection{Ablation Study}
\begin{figure*}[!t]
\centering
\includegraphics[width=.9\linewidth]{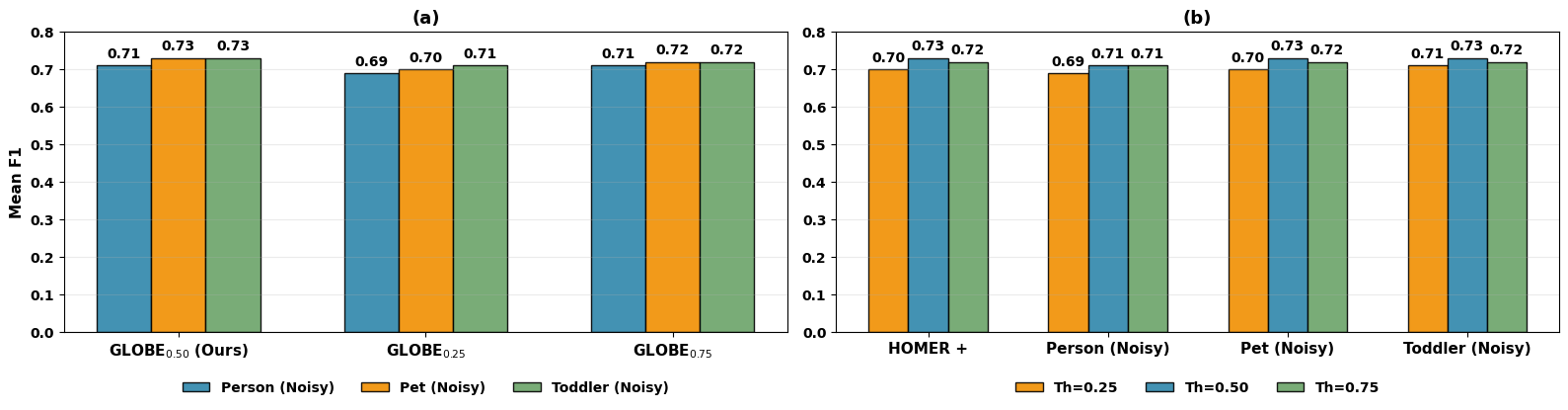}
\caption{ (a) Comparison of mean F1 score under varying noise levels across different noise sources (person, pet, and toddler).
(b) Comparison of mean F1 score across datasets under varying confidence thresholds (0.25, 0.50 and 0.75).}
\label{fig:ablation_study}
\end{figure*}

\noindent
To evaluate the robustness under increasing environmental uncertainty, we analyze model performance across varying levels of structured noise and confidence thresholds. We first consider three noise settings: low noise (0.25), the proposed setting (0.5), and high noise (1.0) in the training set, while the test set still contains the noise level of 1.0. We then examine the effect of different confidence thresholds, comparing a low threshold (0.25), the proposed value (0.5), and a high threshold (0.75), where LLM-based reasoning is triggered more frequently.
\subsubsection{Noise Levels}

As shown in Fig.~\ref{fig:ablation_study}, reducing the noise level to 0.25 slightly decreases the mean F1 score by margins of 0.02, 0.03, and 0.02 for the three noise types, respectively, when compared to the base model with a 0.5 noise level. This could be because of a significant difference in the training and test sets, since the test set contains a noise level of 1.0. In contrast, increasing the noise level to 0.75 keeps the performance similar to the base model, with only a slight difference of 0.01 in Pet and Toddler noise conditions. This could be that the noise level is similar to the test set, but a higher noise level in the training set might cause slight degradation in the learned model predictions. Overall, the variations in the F1 scores across different noise levels is minimal, demonstrating the robustness of our model under various level of noise.


\subsubsection{Confidence Threshold}

Fig.~\ref{fig:ablation_study} shows results for varying the confidence threshold $d$ on the non-noisy HOMER+ and HOMER-Noise with Person, Pet, and Toddler noise types. Across all settings, performance is largely stable with respect to the threshold choice. The best results are consistently obtained with $d=0.5$, while a higher threshold of 0.75 leads to a slight degradation, and a lower threshold of 0.25 results in a more pronounced drop in performance. These results indicate that the framework is not highly sensitive to LLM behavior; however, overly conservative reliance on LLM-based refinement reduces the mean F1 score, highlighting the importance of balancing sequential modeling with selective LLM intervention.





\subsection{Robot Validation}
We validate the proposed framework using a representative scene illustrated in Fig.~\ref{fig:overview_framework}, where the robot observes a household environment containing multiple objects. The platform consists of a Stretch 3 robot equipped with an RGB camera for object identification, while activity labels are provided externally. 
During the observation phase, the robot collects activity and object usage data from the user. As shown in Fig.~\ref{fig:overview_framework}, the robot observes that the user typically drinks coffee while working at a computer around 6 a.m. These observations are used to train the prediction model, enabling it to learn associations between activities and object usage. In the assistance phase, the model predicts both the next activity and the corresponding objects likely to be used. Before acting, the robot evaluates the prediction confidence: if sufficiently high, it performs proactive assistance. In the illustrated example, the robot predicts the need for the coffee object at 6 a.m. and delivers it to the user. When prediction confidence is low, the LLM is invoked to refine the estimate. The robot continues to observe user behavior and incrementally updates the model with newly observed activity–object pairs, allowing prediction performance to improve over time as additional data are collected.






%% file: sections/conclusion.tex
\section{Conclusions}
\label{sec:conclusion}
\noindent
In this paper, we address the problem of proactive robot assistance for routine object usage prediction in dynamic household environments, where human behavior is inherently variable, and observations are often noisy. 
We propose GLOBE, a lightweight framework that combines n-gram Markov models for capturing temporal behavioral patterns with 
LLM-based reasoning to handle uncertainty. 
To support evaluation under realistic conditions, we introduced HOMER-Noise, a noisy extension of the HOMER+ dataset that simulates structured environmental disturbances such as object movements caused by humans, pets, and toddlers. Experimental results on both HOMER+ and HOMER-Noise demonstrate that our approach achieves competitive performance with SOTA methods 
while providing improved robustness 
and computational efficiency during model training. 
We hope that this work can lead to developing more robust models for personalized robotic assistance in household environments. 

In future work, we plan to conduct an HRI study with human participants interacting with the Stretch 3 mobile manipulator to demonstrate real-time learning with diverse users
, 
and integrate real-time activity recognition 
with compact, quantized open-source LLMs locally on the robot to reduce inference latency.

%% file: references.bib
@inproceedings{ayub2024interactive,
  title={Interactive continual learning architecture for long-term personalization of home service robots},
  author={Ayub, Ali and Nehaniv, Chrystopher L and Dautenhahn, Kerstin},
  booktitle={ICRA},
  pages={11289--11296},
  year={2024},
  organization={IEEE}
}

@inproceedings{puig2021watchandhelp,
    title={{Watch-And-Help: A Challenge for Social Perception and Human-AI Collaboration}},
    author={Xavier Puig and Tianmin Shu and Shuang Li and Zilin Wang and Yuan-Hong Liao and Joshua B. Tenenbaum and Sanja Fidler and Antonio Torralba},
    booktitle={ICLR},
    year={2021},
}

@inproceedings{patel2022proactive,
  title={Proactive Robot Assistance via Spatio-Temporal Object Modeling},
  author={Patel, Maithili and Chernova, Sonia},
  booktitle={CoRL},
  year={2022}
}

@inproceedings{patel2023predicting,
  title={Predicting Routine Object Usage for Proactive Robot Assistance},
  author={Patel, Maithili and Prakash, Aswin and Chernova, Sonia},
  booktitle={CoRL},
  year={2023}
}

@inproceedings{gross_robot_2015,
	address = {Hamburg, Germany},
	title = {Robot companion for domestic health assistance: {Implementation}, test and case study under everyday conditions in private apartments},
	isbn = {978-1-4799-9994-1},
	shorttitle = {Robot companion for domestic health assistance},
	doi = {10.1109/IROS.2015.7354230},
	urldate = {2026-03-25},
	booktitle = {IROS},
	publisher = {IEEE},
	author = {Gross, Horst-Michael and Mueller, Steffen and Schroeter, Christof and Volkhardt, Michael and Scheidig, Andrea and Debes, Klaus and Richter, Katja and Doering, Nicola},
	month = sep,
	year = {2015},
	pages = {5992--5999},
}

@article{alaghbari_activities_2022,
	title = {Activities {Recognition}, {Anomaly} {Detection} and {Next} {Activity} {Prediction} {Based} on {Neural} {Networks} in {Smart} {Homes}},
	volume = {10},
	copyright = {https://creativecommons.org/licenses/by/4.0/legalcode},
	issn = {2169-3536},
	doi = {10.1109/ACCESS.2022.3157726},
	urldate = {2026-03-25},
	journal = {IEEE Access},
	author = {Alaghbari, Khaled A. and Md. Saad, Mohamad Hanif and Hussain, Aini and Alam, Muhammad Raisul},
	year = {2022},
	pages = {28219--28232},
}

@article{ayub_continual_2025,
	title = {Continual {Learning} {Through} {Human}-{Robot} {Interaction}: {Human} {Perceptions} of a {Continual} {Learning} {Robot} in {Repeated} {Interactions}},
	volume = {17},
	issn = {1875-4791, 1875-4805},
	shorttitle = {Continual {Learning} {Through} {Human}-{Robot} {Interaction}},
	doi = {10.1007/s12369-025-01214-9},
	language = {en},
	number = {2},
	urldate = {2026-03-25},
	journal = {International Journal of Social Robotics},
	author = {Ayub, Ali and De Francesco, Zachary and Holthaus, Patrick and Nehaniv, Chrystopher L. and Dautenhahn, Kerstin},
	month = feb,
	year = {2025},
	pages = {277--296},
}

@article{markov_example_2006,
	title = {An {Example} of {Statistical} {Investigation} of the {Text} \textit{{Eugene} {Onegin}} {Concerning} the {Connection} of {Samples} in {Chains}},
	volume = {19},
	issn = {0269-8897, 1474-0664},
	doi = {10.1017/S0269889706001074},
	language = {en},
	number = {4},
	urldate = {2026-03-26},
	journal = {Science in Context},
	author = {Markov, A. A.},
	month = dec,
	year = {2006},
	pages = {591--600},
}

@inproceedings{mainprice_human-robot_2013,
	address = {Tokyo},
	title = {Human-robot collaborative manipulation planning using early prediction of human motion},
	isbn = {978-1-4673-6358-7 978-1-4673-6357-0},
	doi = {10.1109/IROS.2013.6696368},
	urldate = {2026-03-26},
	booktitle = {IROS},
	publisher = {IEEE},
	author = {Mainprice, Jim and Berenson, Dmitry},
	month = nov,
	year = {2013},
	pages = {299--306},
}

@inproceedings{hawkins_anticipating_2014,
	address = {Hong Kong, China},
	title = {Anticipating human actions for collaboration in the presence of task and sensor uncertainty},
	isbn = {978-1-4799-3685-4},
	doi = {10.1109/ICRA.2014.6907165},
	urldate = {2026-03-26},
	booktitle = {ICRA},
	publisher = {IEEE},
	author = {Hawkins, Kelsey P. and Bansal, Shray and Vo, Nam N. and Bobick, Aaron F.},
	month = may,
	year = {2014},
	pages = {2215--2222},
}

@inproceedings{cuntoor_human-robot_2012,
	address = {Vilamoura-Algarve, Portugal},
	title = {Human-robot teamwork using activity recognition and human instruction},
	isbn = {978-1-4673-1736-8 978-1-4673-1737-5 978-1-4673-1735-1},
	doi = {10.1109/IROS.2012.6385698},
	urldate = {2026-03-26},
	booktitle = {IROS},
	publisher = {IEEE},
	author = {Cuntoor, Naresh P. and Collins, Roderic and Hoogs, Anthony J.},
	month = oct,
	year = {2012},
	pages = {459--465},
}

@inproceedings{wang_lami_2024,
	address = {Honolulu HI USA},
	title = {{LaMI}: {Large} {Language} {Models} for {Multi}-{Modal} {Human}-{Robot} {Interaction}},
	isbn = {979-8-4007-0331-7},
	shorttitle = {{LaMI}},
	doi = {10.1145/3613905.3651029},
	language = {en},
	urldate = {2026-02-22},
	booktitle = {Extended {Abstracts} of the {CHI} {Conference} on {Human} {Factors} in {Computing} {Systems}},
	publisher = {ACM},
	author = {Wang, Chao and Hasler, Stephan and Tanneberg, Daniel and Ocker, Felix and Joublin, Frank and Ceravola, Antonello and Deigmoeller, Joerg and Gienger, Michael},
	month = may,
	year = {2024},
	pages = {1--10},
}

@inproceedings{ren_robots_2023,
    title={Robots That Ask For Help: Uncertainty Alignment for Large Language Model Planners},
	author = {Ren, Allen Z. and Dixit, Anushri and Bodrova, Alexandra and Singh, Sumeet and Tu, Stephen and Brown, Noah and Xu, Peng and Takayama, Leila and Xia, Fei and Varley, Jake and Xu, Zhenjia and Sadigh, Dorsa and Zeng, Andy and Majumdar, Anirudha},
    booktitle={CoRL},
    year={2023}
}

@inproceedings{sun_assistantx_2025,
	address = {Hangzhou, China},
	title = {{AssistantX}: {An} {LLM}-{Powered} {Proactive} {Assistant} in {Collaborative} {Human}-{Populated} {Environments}},
	copyright = {https://doi.org/10.15223/policy-029},
	isbn = {979-8-3315-4393-8},
	shorttitle = {{AssistantX}},
	doi = {10.1109/IROS60139.2025.11246901},
	urldate = {2026-03-26},
	booktitle = {IROS},
	publisher = {IEEE},
	author = {Sun, Nan and Mao, Bo and Li, Yongchang and Guo, Di and Liu, Huaping},
	month = oct,
	year = {2025},
	pages = {3352--3359},
}
